# Semantic Segmentation Using Super Resolution Technique as Pre-Processing

Chih-Chia Chen[1], Wei-Han Chen[1], Jen-Shiun Chiang[1*], Chun-Tse Chien[1] and Tingkai Chang[2]

[1]Department of Electrical and Computer Engineering, Tamkang University
[2]Crean Lutheran High School, Irvine, CA 92618, USA
[*]Corresponding author(s). E-mail(s): chiang@mail.tku.edu.tw
[1]Contributing authors: crystal88irene@gmail.com; jkj211378@gmail.com; popper0927@hotmail.com
[2]tingkaichang1@gmail.com

*Abstract*—Combining high-level and low-level visual tasks is a common technique in the field of computer vision. This work integrates the technique of image super resolution to semantic segmentation for document image binarization. It demonstrates that using image super-resolution as a preprocessing step can effectively enhance the results and performance of semantic segmentation.

## I. INTRODUCTION

Combining semantic segmentation and super-resolution can bring numerous benefits. Firstly, integrating semantic segmentation with super-resolution can generate high-resolution images with enhanced semantic information in various applications, such as remote sensing and surveillance applications [1], converting low-resolution images into high-resolution ones to find more detail information, and aiding object recognition and scene analysis. Secondly, traditional super-resolution methods may result in smooth images, leading to the loss of fine details. However, by leveraging semantic segmentation, it becomes possible to identify which regions should have higher resolution, and therefore it can preserve the intricate details.

## II. RELATED WORK

### A. Super Resolution

Single-image super-resolution is an inverse problem aiming at restoring a corresponding high-resolution image from a low-resolution input. Deep learning-based super-resolution methods have been widely proposed and have achieved state-of-the-art (SOTA) performance in various benchmark tests due to their data and context adaptability. Based on machine learning approaches, there are two main architectures for super-resolution: (1) CNN-based SR methods: the first application of CNN to single-image super-resolution was introduced by SRCNN [2], which made significant progress compared to traditional methods, such as Bicubic interpolation. (2) Transformer-based models have gained increasing attention in recent years, as it can model long-term dependence of image. Many Transformer-based methods have been proposed for computer vision tasks. For example, SwinIR [3] directly

This paper is supported by Nation Science and Technology Council, under grant number: 111-2221-E-032-021-.

applies the Swin Transformer to image restoration tasks and achieves excellent SOTA results.

### B. Semantic Segmentation

This work aims to classify and binarize the text and background of document images using semantic segmentation techniques while preserving the readability of the document. Tensmeyer *et al.* [4] combined P-FM and FM losses to train FCN for the binarization task of document images. Ronneberger *et al.* [5] proposed U-Net, a U-shaped network architecture, to capture contextual and positional information, and utilize shortcut connections to retain spatial information lost during subsampling operations.

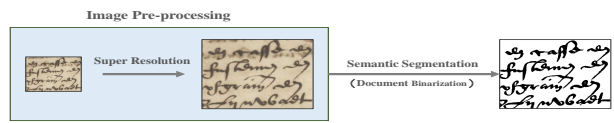

Fig. 1. Overall flow chart of the two-stage framework.

## III. PROPOSED METHOD

In this paper, we propose a framework that combines super-resolution model with semantic segmantation. As depicted in Fig. 1, our framework comprises two stages, where the first stage uses the SwinOIR [6] algorithm model as pre-processing, and the second stage performs DocEnTr [7] as semantic segmantation model on the high resolution images.

To demonstrate the impact of image super-resolution as a preprocessing step, the following experimental setup was conducted. Firstly, the ground truth and original document images from the used binarization dataset were resized to half of their original size. Subsequently, the images were directly resized to half of the size of the original document images for semantic segmentation processing. The output results were then compared with the downsized ground truth, referred to as "w/o SR" (without super-resolution). As for the experimental control group, the original document images from the dataset were resized to half of their original size. The super-resolution technique was applied to enlarge them by a factor of 2. Then, semantic segmentation was performed on the upscaled images, and the output results were compared with the ground truth from the original dataset, referred to as "w/ SR" (with super-resolution).

## IV. EXPERIMENTS

### A. Dataset

DIBCO (Document Image Binarization Contest) is a public dataset provided by the Document Image Binarization Contest. It consists of nine datasets that include grayscale, color machine-printed, and handwritten images. In this research work, we employed two of these datasets, namely DIBCO 2017 [8] and H-DIBCO 2018 [9].

### B. Experimental Results

From Table I, it can be observed that using super-resolution as an image preprocessing step yields better results in both datasets. In the DIBCO 2017 [8] dataset, the results with super-resolution preprocessing show a PSNR improvement of 1.8 dB compared to the results without super-resolution. In the H-DIBCO 2018 [9] dataset, the improvement is even around 4 dB, and the SSIM values also show an improvement of approximately 0.05 and 0.02, respectively. This work also provides the visual comparison in Figures 2 and 3.

The results of "w/ SR" exhibit a noticeable increase in black block-like regions and thicker text compared to the ground truth (GT). This is likely due to the fact that semantic segmentation is a pixel-level computer vision task, and with a lower resolution, there is a higher chance of mistaking dirt for text, leading to a decline in performance. On the other hand, the super-resolution task excels at producing clearer boundaries for lines and colors in the image. By increasing the pixel count, it enables finer discrimination of text, dirt, and background.

TABLE I
COMPARISON TABLE OF PRE-PROCESSING VALUES FOR SEMANTIC SEGMENTATION USING IMAGE SUPER-RESOLUTION

| Method | DIBCO 2017 [8] | | H-DIBCO 2018 [9] | |
|--------|------|------|------|------|
|  | PSNR | SSIM | PSNR | SSIM |
| w/ SR | 44.44 | 0.9341 | 27.13 | 0.4919 |
| w/o SR | 42.62 | 0.8827 | 23.81 | 0.4758 |

## V. CONCLUSIONS

The results of this study demonstrate that incorporating super-resolution as an image preprocessing step significantly improves the performance and results in the semantic segmentation task of document images. It enables finer discrimination of text, dirt, and background, which is crucial for enhancing the accuracy and readability of document image processing. The future research can further explore the combination of different super-resolution and semantic segmentation methods, as well as their application in other relevant computer vision tasks. Additionally, the use of more datasets and evaluation metrics would contribute to a more comprehensive assessment of the method's performance across different contexts. These efforts will contribute to the advancement and innovation of document image processing and analysis field.

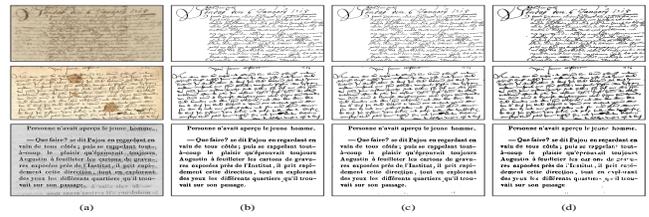

Fig. 2. Bizarization results of three example images in DIBCO 2017: (a) the original input image, (b) the ground truth image, (c) 'w/SR' binarization output image, (d) 'w/o SR' binarization output image

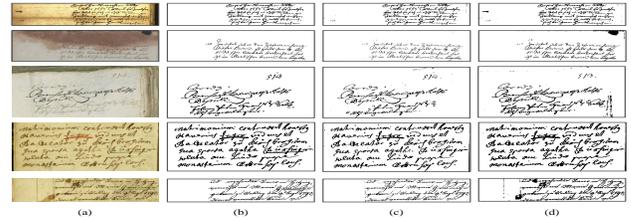

Fig. 3. Bizarization results of three example images in HDIBCO 2017: (a) the original input image, (b) the ground truth image, (c) 'w/SR' binarization output image, (d) 'w/o SR' binarization output image